\documentclass{article}

\usepackage{arxiv}

\usepackage[utf8]{inputenc} 
\usepackage[T1]{fontenc}    
\usepackage{hyperref}       
\usepackage{url}            
\usepackage{booktabs}       
\usepackage{amsfonts}       
\usepackage{nicefrac}       
\usepackage{microtype}      
\usepackage{lipsum}
\usepackage{comment}
\usepackage{svg}

\title{Ethically aligned Deep Learning: Unbiased Facial Aesthetic Prediction }
\author{
  Michael~Danner\footnote[1]~~~\footnote[2]{Reutlingen}~
  , Thomas Weber\footnotemark[2]~
  , Leping~Peng\footnote[3]{Hunan}~
  , Tobias Gerlach\footnotemark[2]~
  , Xueping Su\footnote[4]{Xian}~
  , Matthias Rätsch\footnotemark[2]~~\footnotemark[4]
  \AND
  \\
  \footnotemark[1]~ University of Surrey
  \footnotemark[2]~ Reutlingen University
  \footnotemark[3]~ Hunan University of Science and Technology\\
  \footnotemark[4]~ Xi’an Polytechnic University
}

\begin{document}
\maketitle

\begin{abstract}

Facial beauty prediction (FBP) aims to develop a machine that automatically makes facial attractiveness assessment.
In the past those results were highly correlated with human ratings, therefore also with their bias in annotating.
As artificial intelligence can have racist and discriminatory tendencies, the cause of skews in the data must be identified. 
Development of training data and AI algorithms that are robust against biased information is a new challenge for scientists.
As aesthetic judgement usually is biased, we want to take it one step further and propose an Unbiased Convolutional Neural Network for FBP.
While it is possible to create network models that can rate attractiveness of faces on a high level, from an ethical point of view, it is equally important to make sure the model is unbiased.
In this work, we introduce AestheticNet, a state-of-the-art attractiveness prediction network, which significantly outperforms competitors with a Pearson Correlation of 0.9601.
Additionally, we propose a new approach for generating a bias-free CNN to improve fairness in machine learning.
\end{abstract}

%



 \section{Introduction}
In 2016 "Beauty.ai", a Hong-Kong based technology company hosted the first international beauty contest judged by artificial intelligence \cite{beautyai} but the results are heavily biased for example against dark skin \cite{guardian_beautyai}.
``Machine learning models are prone to biased decisions, due to biases in data-sets''\cite{sharma_data_2020}. Biased training data potentially leads to discriminatory models, as the data-sets are created by humans or derived from human activities in the past, for example hiring algorithms \cite{bogen_all_2019}.
The reason for racist and discriminatory tendencies must be identified.
As the learning algorithms become more complex, understanding why the decisions are made, or even how, prove to be nearly impossible \cite{bostrom_ethics_2018}. 
Therefore the development of non-biased training data and AI algorithms (defined by the European Commission High-Level Expert Group on Artificial Intelligence \cite{ai_def}) is a new and increasingly complex challenge for scientists around the world. 
The specific field of aesthetic judgement is especially vulnerable to being biased, as aesthetic judgement itself is already a subjective rating \cite{richmond_beholders_2017}.
Our data analysis has already proven that people consider their own ethnicity to be more attractive than others \cite{gerlach_who_2020}, this is the biggest bias in our research and within our data. 
With this tendency, it becomes difficult to generate data to train a machine-learning algorithm, which assesses a person's attractiveness without bias.

\section{State of the art} 
\label{sec:state-of-the-art}

Machine learning has evolved in the past decades and stands out due to the fact that the knowledge in the system is not provided by experts. 
Facial Beauty Prediction (FBP), that is consistent with human perception, is a significant visual recognition problem and a much-studied subject in recent decades. 
Eisenthal et al. \cite{eisenthal_facial_2006} and Kagian et al. \cite{kagian_machine_2008} were among the first to publish their research about automatic facial attractiveness predictors and supervised learning techniques, based on the extraction of feature landmarks on faces.

With the introduction of convolutional neural networks (CNN) and large-scale image repositories, facial image and video tasks get more powerful~\cite{krizhevsky_imagenet_2017, zeiler_visualizing_2013, deng_imagenet_2009}.
Xie~et~al.~\cite{scut-fbp500} present the SCUT-FBP500 dataset, containing 500 asian female subjects with attractiveness ratings. 
Since ``FBP is a multi-paradigm computation problem'' the successor SCUT-FBP5500~\cite{scut-fbp5500_2018} is introduced in 2018, including an increased database of 5500 frontal faces with multiple attributes: male/female, Asian/Caucasian, age, beauty score.
Liang~et~al.~(2018) evaluated their database ``using different combinations of feature and predictor, and various deep learning methods'' on AlexNet~\cite{krizhevsky_imagenet_2017}, ResNet-18~\cite{DBLP:journals/corr/HeZRS15} and ResNeXt-50 and achieved the Pearson Correlation \textit{PC:} $0.8777$; mean average error \textit{MAE:} $0.2518$; root-mean-square error \textit{RMSE:} $0.3325$ as a benchmark.
It was observed, that all deep CNN models are superior to the shallow predictor with hand-crafted geometric feature or appearance feature~\cite{scut-fbp5500_2018}.

\section{Ethics}
Since a large number of studies have shown that machines trained by humans will also repeat human bias \cite{zuiderveen_borgesius_discrimination_2018, stephens-davidowitz_cost_2014, munger_tweetment_2017, horvitz_ai_2017}.
Unfortunately, everyone is biased, because a large part of human bias is unconscious and it is not easy to be noticed, which was demonstrated in a Seattle press conference at the University of Washington by psychologists who developed a new tool that measures the unconscious bias \cite{joel_schwarz_roots_nodate}.
Unconscious bias comes from educational background, culture, attitudes, and stereotypes we pick up from the world we live in.

An example of gender bias: many people have the stereotype that females are worse at mathematics tasks and better at verbal tasks than males~\cite{johns_knowing_2005}.
A Yale study shows that male candidates are judged to be more talented and experienced and hired more often~\cite{moss-racusin_science_2012}.
Affinity bias refers to unconsciously preferring familiar people, this is verified in our first part of our experiment. 
A machine trained on a data-set labelled by Europeans judges European faces higher than Asian faces and vice versa. 
And when you unconsciously notice people’s appearances and associate it with their personality, you might have beauty bias.

How to build a fair artificial intelligence if human beings are biased? 
Many technology companies begin to realise this problem and take efforts, such as Google \cite{google_llc_rework_nodate} and Facebook \cite{facebook_inc_managing_nodate}. 
Facebook makes unconscious bias training videos widely available and Google educates employees by an unconscious bias training program to reduce human bias during human-computer interaction.

Bias also exists in the field of image processing, for example, artificial intelligence applications tag pictures of White American brides as ``brides'', ``dresses'', and ``weddings'' while pictures of North Indian brides are tagged as ``performing arts'' and ``costumes'' \cite{shankar_no_2017}.
Angwin's and Larson's~\cite{larson_bias_2016} analysis of ethical bias has prompted research showing that the disparity can be addressed if the algorithms focus on the fairness of outcomes.

What is the significance of our research? 
Although both Hume \cite{david_standard_1898} and Kant \cite{kant_critique_2000} believe that aesthetic judgement is only a subjective feeling, which regarding the pleasure that we take from a beautiful object, and aesthetics itself is neither right nor wrong nor moral, the behaviour caused by aesthetic judgements are related to morality \cite{cui_different_2019, haidt_emotional_2001}.
Many studies show that unfair recruitment cases encountered due to aesthetic judgements, \cite{maddox_racial_2018, mason_appearance_2017, beattie_possible_2012}, more attractive people have higher incomes than less attractive individuals \cite{anyzova_beauty_2018, french_physical_2002, cawley_impact_2004}.
Our research provides a practical example of how to build a fair and trustworthy AI.

\section{Experiment}
\label{sec:our-experiment}

We conducted online surveys on image datasets beginning in 2013 on multiple datasets where thousands of students and their relatives participated.
Starting 2017 we used the Asian-European-dataset SCUT-FBP to evaluate biased annotations from Chinese and German universities. 
The results proved the assumption that German students favour images of European women and vice versa Chinese students rate Asian portraits higher. 
Within our research, we use Convolutional Neural Networks (CNN) for facial aesthetic score prediction and introduce AestheticNet as our approach.

Since the SCUT-FBP 5500 dataset is a small dataset for deep learning tasks, we use data augmentation methods to enlarge the sample size of the training set by generating GAN images with either Asian or European or mixed images as input and new synthesised images as output. This augmentation method proves superior to geometric transformations like cropping and rotating. All images are preprocessed by normalisation methods to harmonise face pose, facial landmark positions and image size. For our training data, we use the SCUT-FBP dataset and labels enhanced with Chinese annotators to train an Asian-biased network first. The second network is trained with the same dataset but labelled by German students and their relatives.

We compare our method with other state-of-the-art approaches on the SCUT-FBP500 datasets. As shown in Table~\ref{tab:comparison} our AestheticNet therefore significantly surpasses previous approaches, which is mainly due to the augmentation with synthetic images and the optimisation of the previous approaches. In our best experiment, we achieve a Pearson correlation of 0.9601, a normalised mean average error of 3.896\% and a normalised root mean squared error of 5.580\%. The results are normalised because there are different datasets with different score ranges.

\begin{table}[]
    \centering
    \caption{Comparison of prediction accuracy on SCUT-FBP5500}
    \begin{tabular}{l|c c c}
        \toprule
                       & PC              & nMAE (\%)      & nRMSE (\%)     \\ \hline
        AlexNet~\cite{scut-fbp5500_2018} & 0.8298          & 7.345          & 9.548          \\ 
        AlexNet~\cite{DBLP:journals/cin/ZhaiCDGPZ19} & 0.8634          &                &                \\ 
        ResNet-18~\cite{scut-fbp5500_2018}      & 0.8513          & 7.045          & 9.258          \\ 
        ResNeXt-50~\cite{scut-fbp5500_2018}     & 0.8777          & 6.295          & 8.313          \\ 
        HMTNet\cite{xu_hierarchical_2019} & 0.8783 & 6.2525 &  \\ 8.158
        AaNet~\cite{DBLP:conf/ijcai/LinLJC19}       & 0.9055          & 5.590          & 7.385          \\ 
        P-AaNet~\cite{DBLP:conf/ijcai/LinLJC19}        & 0.8965          & 5.713          & 7.588          \\ 
        2M BeautyNet~\cite{DBLP:journals/access/GanSXZMHZBLP20}   & 0.8996          &                &                \\
        \hline
        EfficientNetB3 based AestheticNet (ours) & 0.9011          & 5.841          & 7.663          \\ 
        VGG-Face based AestheticNet (ours)       & 0.9363          & 4.400          & 6.261          \\ 
        \hline
        \textbf{AestheticNet (ours)}  & \textbf{0.9601} & \textbf{3.896} & \textbf{5.580} \\ 
        \bottomrule
    \end{tabular}
    \label{tab:comparison}
\end{table}


\begin{figure}[t]
	\centering
	\includegraphics[width=0.8\textwidth]{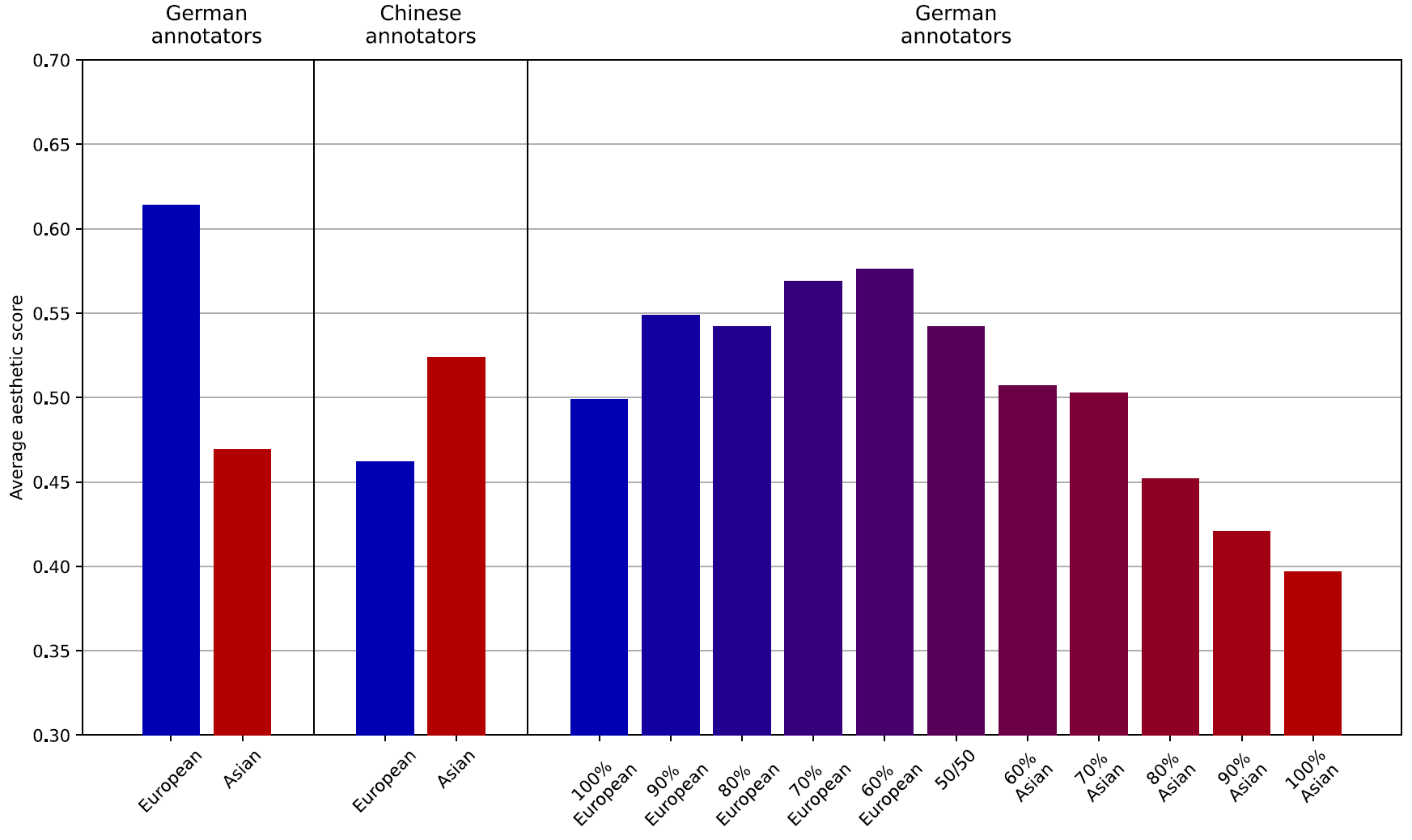}
	\caption{Unconscious bias towards ethnic aesthetic of either German or Chinese annotators. Left: average aesthetic score on SCUT-FBP by German annotators, middle: average aesthetic score labelled by Chinese students, right: aesthetic scores on the Eurasian dataset annotated by German students.}
	\label{fig:figure_asian}
\end{figure}

\begin{figure}[t]
	\centering
	\includegraphics[width=0.8\textwidth]{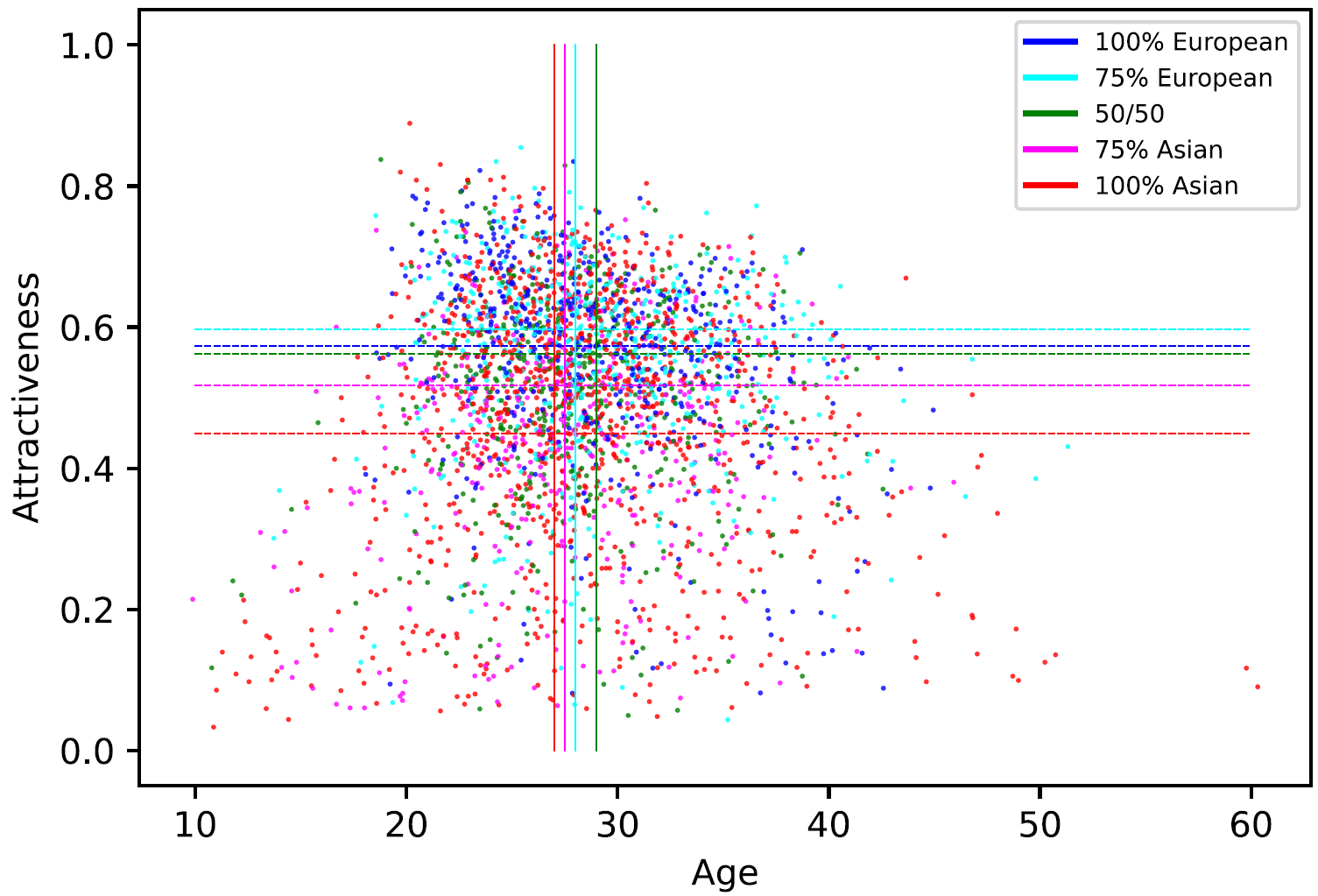}
	\caption{Biased correlation between attractiveness, age and ethnicity by German annotators. In an ethnical fair network the attractiveness would be the same height for equal age groups.}
	\label{fig:figure_bubbles}
\end{figure}

Figure~\ref{fig:figure_asian} shows the biased average score of our networks on the SCUT-FBP dataset and the Eurasian dataset. 
Figure~\ref{fig:figure_bubbles} illustrates the analysis on the distribution of aesthetic score and age for Asians, Europeans and three mixed-racial subgroups. In a fair machine, the distance of all lines would overlap themselves.

Having a state-of-the-art aesthetic prediction network, we then train a third CNN on the features from the Asian and German labelled networks to generate a non-biased network. Therefore, the synthesised Eurasian dataset is used with a categorical-cross-entropy-loss-function to converge the subgroup's intersection points of Figure~\ref{fig:figure_bubbles}. 

\section{Conclusion}
Our two main contributions are AestheticNet and a new approach to bias-free machine learning tools.
In this work, we have proposed to augment the SCUT-FBP dataset by synthesised GAN images and show that AestheticNet predicts facial attractiveness with higher correlation then competitive approaches. Then we utilise categorical-cross-entropy-loss learning strategies to minimise bias in networks. 
Unbiased networks are an important step towards a future, where more decisions are made by AI and therefore more lives are influenced by artificial intelligence - unbiased decision making is the foundation of ethical and moral values.

Bias-free decision making is a challenging problem in machine learning tasks, yet it yields the great potential to be one of the most significant strengths of an AI. We have shown a method to eliminate bias in facial attractiveness prediction and this method can be transferred to multiple similar networks.

\bibliography{main} 
\bibliographystyle{ieeetr}

\end{document}